\def\year{2020}\relax
\documentclass[letterpaper]{article} 
\usepackage{aaai20}  
\usepackage{times}  
\usepackage{helvet} 
\usepackage{courier}  
\usepackage[hyphens]{url}  
\usepackage{graphicx} 
\usepackage{graphicx}  
\usepackage[utf8]{inputenc} 
\usepackage[T1]{fontenc}    
\usepackage{hyperref}       
\usepackage{url}            
\usepackage{booktabs}       
\usepackage{amsfonts}       
\usepackage{amssymb}
\usepackage{amsmath,amsthm}
\usepackage{subcaption}
\usepackage{algpseudocode}
\usepackage{algorithm}
\usepackage{array,multirow}
\usepackage{nicefrac}       
\usepackage{microtype}      
\usepackage{xcolor}
\usepackage{graphicx}
\usepackage{wrapfig}

\newtheorem{theorem}{Theorem}
\newtheorem*{theorem*}{Theorem}

\usepackage[normalem]{ulem}

\newcommand{\eat}[1]{}

\title{Lifted Hybrid Variational Inference}
\author{Yuqiao Chen\thanks{Equal contribution} , Yibo Yang\footnotemark[1] , Sriraam Natarajan , Nicholas Ruozzi\\
The University of Texas at Dallas\\ 
800 W Campbell Rd, Richardson, Texas, USA\\
\{yuqiao.chen, yibo.yang, sriraam.natarajan, nicholas.ruozzi\}@utdallas.edu
}
 
\begin{document}

\maketitle

\begin{abstract}
A variety of lifted inference algorithms, which exploit model symmetry to reduce computational cost, have been proposed to render inference tractable in probabilistic relational models.  Most existing lifted inference algorithms operate only over discrete domains or continuous domains with restricted potential functions, e.g., Gaussian. We investigate two approximate lifted variational approaches that are applicable to hybrid domains and expressive enough to capture multi-modality.
We demonstrate that the proposed variational methods are both scalable and can take advantage of approximate model symmetries, even in the presence of a large amount of continuous evidence. We demonstrate that our approach compares favorably against existing message-passing based approaches in a variety of settings. Finally, we present a sufficient condition for the Bethe approximation to yield a non-trivial estimate over the marginal polytope.
\end{abstract}

\section{Introduction}

Lifted methods have recently gained popularity due to their ability to handle previously intractable probabilistic inference queries over Markov random fields (MRFs) and their generalizations.  These approaches work by exploiting symmetries in the given model construct groups of indistinguishable random variables that can be used to collapse the model into a simpler one on which inference is more tractable. 

High-level approaches to lifted inference include message-passing algorithms such as lifted belief propagation \cite{singla2008,kersting2009} and lifted variational methods \cite{bui2014,gallo2018}.  
The common theme across these methods is the construction of a lifted graph on which the corresponding inference algorithms are run.  The message-passing algorithms are applied directly on the lifted graph, while lifted variational methods encode symmetries in the model as equality constraints in the variational objective.  These two approaches are directly related via the same variational objective, known as the Bethe free energy \cite{yedidia2001,yedweiss}. While successful, these methods were designed for discrete relational MRFs.

Existing work on lifting with continuous domains has focused primarily on Gaussian graphical models \cite{choi2010,choi2011,ahmadi2011}. Other lifted inference methods for generic hybrid domains \cite{choi2012} use expectation maximization to learn variational approximations of the MRF potentials (which requires sampling from the potentials, hence implicitly assumes the potentials are normalizable) and then perform lifted variable elimination or MCMC on the resulting variational model.  While applicable to generic models, this approach is somewhat complex and can be expensive on lifted graphs with large treewidth.

Our aim is to provide a general framework for lifted variational inference that can be applied to both continuous and discrete domains with generic potential functions. Our approach is based on mixtures of mean-field models and a choice of entropy approximation.  We consider two entropy approximations, one based on the Bethe free energy, whose local optima are closely related to fixed points of BP \cite{yedidia2001}, and a lower bound on the differential entropy based on Jensen's inequality \cite{gershman2012}.

We make the following key contributions in this work: (1) We develop a general lifted hybrid variational approach for probabilistic inference. (2) We consider two different types of approximations based on mixtures of mean-field models.  To our knowledge a systematic comparison of these two different approximations for continuous models does not exist in the literature. (3) We provide theoretical justification for the Bethe free energy in the continuous case by providing a sufficient condition for it to be bounded from below over the marginal polytope. (4) We demonstrate the superiority of our approach empirically against particle-based message-passing algorithms and variational mean field.  A key attribute of our work is that it {\em does not make any distributional or model assumptions} and can be applied to arbitrary factor graphs.

\section{Preliminaries}

Given a hypergraph $G=(\mathcal{V}, \mathcal{C} )$ with node set $\mathcal{V}$ and hyper-edge/clique set $\mathcal{C}\subseteq 2^\mathcal{V}$, such that each node $i \in \mathcal{V}$ corresponds to a random variable $x_i \in \mathcal{X}_i$, a Markov random field (MRF) defines a joint probability distribution
\begin{align}
p(x) = \frac{1}{\mathcal{Z}} \prod_{c \in \mathcal{C}} \psi_c(x_c),\label{eq:dist}
\end{align}
where each $\psi_c: \prod_{i \in c} \mathcal{X}_i \to \mathbb{R}_{\geq 0}$ is a non-negative potential function associated with a clique $c$ in $\mathcal{C}$, and $\mathcal{Z}$ is a normalizing constant that ensures $p$ is a probability density.  In this work, we consider hybrid MRFs that may contain both discrete and continuous random variables, so that $\mathcal{X}_i$ may either be finite or uncountable.  For example, if all of the variables have continuous domains and the product of potential functions is integrable, then the normalization constant exists, e.g., $\mathcal{Z} \triangleq \int_{x\in \prod_{i\in \mathcal{V}}\mathcal{X}_i} \prod_{c \in \mathcal{C}} \psi_c(x_c) < \infty$.  The hypergraph $G$ is often visualized as a factor graph that has vertices for both the cliques/factors and the variables, with an edge joining the factor node for clique $c$ to the variable node $i$ if $i\in c$.

We consider two probabilistic inference tasks for a given MRF: (1) marginal inference, i.e., computing the marginal probability distribution $p(x_\mathcal{A})$ of a set of variables $\mathcal{A} \subseteq {\mathcal{V}}$, a special case of which is computing the partition function $\mathcal{Z}$ when $\mathcal{A} = \mathcal{V}$ and (2) maximum a posteriori (MAP) inference, i.e., computing a mode $\arg \max_x p(x_\mathcal{A})$ of the distribution $p(x_\mathcal{A})$. In many applications, we will be given observed values $\hat x_\mathcal{B}$ for a set of variables $\mathcal{B} \subseteq \mathcal{V}$, and the corresponding conditional marginal / MAP inference tasks involve computing $p(x_\mathcal{A} | \hat  x_\mathcal{B})$ instead of $p(x_\mathcal{A})$.

\subsection{Variational Inference}
Variational inference (VI) solves the inference problem approximately by minimizing some divergence measure $D$, often chosen to be the Kullback-Leibler divergence, between distributions of the form \eqref{eq:dist} and a family of more tractable approximating distributions $\mathcal{Q}$, to obtain a surrogate distribution $q^* \in \arg \min_{q \in \mathcal{Q}} D(q, p)$.

The set $\mathcal{Q}$ is typically chosen to trade-off between the computational ease of inference in a surrogate model $q$ and its ability to model complex distributions. When $D$ is the KL divergence, the optimization problem is equivalent to minimizing the variatinoal free energy.
\begin{align*}
\mathcal{F}(q) &:= D_{KL}(q\|p) - \log \mathcal{Z}\\ 
&=  - \sum_{c\in\mathcal{C}} \mathbb{E}_q[\psi_c] - \mathbb{H}[q],
\end{align*}
where $\mathbb{H}[q]$ denotes the entropy of the distribution $q$.  Assuming one can find a $q^*\in{\arg\min}_q \mathcal{F}(q)$, the \textit{simpler} model $q^*$ can be used as a surrogate for inference.  One of the most common choices for $\mathcal{Q}$ is the mean-field approximation in which $\mathcal{Q}$ is selected to be the set of completely factorized distributions.  This choice is popular as the optimization problem is relatively easy to solve for distributions of this form.

\subsection{Lifted Probabilistic Inference via Color Passing}
Lifted inference exploits symmetries that exist in the MRF in order to reduce the complexity of inference.   This is typically done by grouping symmetric variables or cliques together into a single super variable/clique and then tying together the corresponding marginals of all variables in the same super variable/clique \cite{bui2014}. Detecting symmetries can be done by using either a top-down \cite{singla2008} or bottom-up \cite{kersting2009} approach.  We use the color passing (CP) algorithm \cite{kersting2009}, a bottom-up approach that can be applied to arbitrary MRFs.  In CP, all variable and factor nodes are initially clustered based on domain/evidence and the potential functions. Variables with the same domain or with the same evidence value $v$ will be assigned a same color. Each clique node stacks the color of its neighboring nodes in order, appends its own color, and forms a new color. Each variable node collects its neighboring cliques' colors and is assigned a new color. The process is repeated until the colors converge. The color information can be considered as neighborhood structure information and grouping nodes with the same color can be used to compress the graph.  In this work, we use the notation $\#(\mathfrak{c})$ and $\#(\mathfrak{i})$ to denote the number of factor nodes in a super factor $\mathfrak{c}$ and the number of variable nodes in super variable $\mathfrak{i}$, respectively.

\section{Proposed Approaches}

Our goal is to develop a distribution-independent, model-agnostic hybrid lifted inference algorithm that can operate on an arbitrary factor graph. 
To overcome the severe limitations of unimodal variational distributions, e.g., mean-field models, in the hybrid setting, we choose our approximate family $\mathcal{Q}$ to be a family of mixture distributions, and following \cite{jaakkola1998} and \cite{gershman2012}, we require each mixture component to fully factorize\footnote{Note that this assumption is mainly for efficiency. Our approach does not require this assumption for effectiveness.}. Specifically, 
\begin{align}
q(x) &= \sum_{k=1}^K w_k  q^k(x) = \sum_{k=1}^K w_k \prod_i q_i^k(x_i | \eta_i^k),\label{eq:mix}
\end{align}
where $K$ is the number of mixture components, $w_k\geq 0$ is the weight of the $k^{th}$ mixture (a shared parameter across all marginal distributions), and $\sum_{k=1}^{K} w_k = 1$. Each $q_i^k(x_i) \triangleq q_{i}^{k}(x_{i} ; \eta_{i}^{k})$ is some valid distribution with parameters $\eta_{i}^{k}$, e.g., a Gaussian or Beta distribution in the continuous case and a Categorical distribution in the discrete case.  

\subsection{Entropy Approximations}
Ideally, we would find the appropriate model parameters $\eta$ and $w$ by directly minimizing the KL divergence.  Unfortunately, the computation of the entropy $\mathbb{H}[q]$ is intractable for arbitrary variational distributions of the form \eqref{eq:mix}. A notable exception is the case with $K=1$, which is equivalent to the na\"ive mean-field approximation.  In the general case, we consider two tractable entropy approximations:  one based on the Bethe free energy approximation from statistical physics and one based on Jensen's inequality.

\textbf{The Bethe Entropy:} $\mathbb{H}_{B}$ approximates $\mathbb{H}$ as if the graph $G$ associated with $p$ were tree-structured:
\[
\mathbb{H}_\mathrm{B}[q] \triangleq \sum_{c\in {\cal C}} \mathbb{H}[q_c] + \sum_{i\in {\cal V}} (1 - |nb(i)|) \mathbb{H}[q_i],
\]
where $nb(i) = \{ c \in \mathcal{C} |  i \in c \}$ is the set of cliques that contain node $i$ in their scope. The Bethe free energy (BFE) is then defined as $\mathcal{F}_\mathrm{B}(q) \triangleq -\sum_c \mathbb{E}_q[\psi_c] - \mathbb{H}_B[q].$

The BFE approximation is exact whenever the hypergraph $G$ is acyclic, i.e.,  tree-structured.  While variational methods seek to optimize the variational objective directly, message-passing algorithms such as belief propagation (BP) can also be used to find local optima of the BFE \cite{yedidia2001}.  As these types of message-passing algorithms can suffer from convergence issues, gradient based methods that optimize the variational objective directly are sometimes preferred in practice \cite{welling2001}.

\textbf{Jensen's Inequality: } The non-parametric variational inference (NPVI) approximates the entropy using Jensen's inequality \cite{gershman2012}.
\begin{align}
    \mathbb{H}_J[q] &:= -\sum_k w_k \log \left( \sum_j w_j \int q^k(x) q^j(x) dx \right)\label{eq:jensen}
\end{align}
There are two reasons to prefer the Bethe entropy approximation over \eqref{eq:jensen}.  First, the BFE is exact on trees; for tree-structured models, it's likely to outperform NPVI.  Second, \eqref{eq:jensen} doesn't factorize over the graph, potentially making it less useful in distributed settings.  

Conversely, one advantage of \eqref{eq:jensen} over the Bethe entropy is that it yields a provable lower bound on the partition function assuming exact computation. The Bethe entropy only provably translates into a lower bound on tree structured models or for special classes of potential functions \cite{nips2012,uai2013,ruozzi2017}.  Another known drawback of the BFE approximation is that, in the case of continuous random variables, it need not be bounded from below over the so-called local marginal polytope, a further relaxation of the varitional problem in which the optimization over distributions $q\in\mathcal{Q}$ is replaced by a simpler optimization problem over only marginal distributions that agree on their univariate marginals.  This unboundedness can occur even if $p$ corresponds to a multivariate Gaussian distribution \cite{csekeheskesjair2011}.  This potentially makes the BFE highly undesirable for continuous MRFs in practice.

However, for the optimization problem considered here (over a subset of what is referred to as the marginal polytope), it is known that the BFE is bounded from below for Gaussian $p$. Here, we prove that the BFE will be bounded from below over the marginal polytope for a larger class of probability distributions.
\begin{theorem}
\label{thm:bethe}If there exists a collection of densities $g_{i}\in\mathcal{P}(\mathcal{X}_i)$ for each $i\in V$ such that $\sup_{x\in\mathcal{X}} \frac{p(x)}{\prod_i g_i(x_i)} < \infty$, then $\inf_{q\in \mathcal{P}(\mathcal{X})} -\mathbb{E}_q[\log p] - \mathbb{H}_\mathrm{B}[q] > -\infty$, where $\mathcal{P}(\mathcal{X})$ is the set of all probability densities over $\mathcal{X}\triangleq \prod_{i\in V} \mathcal{X}_i$, i.e., the BFE is bounded from below.
\end{theorem}
A number of natural distributions satisfy the condition of the theorem: mean-field distributions, multivariate Gaussian distributions and their mixtures,  bounded densities with compact support, etc.  The proof of this sufficient condition constructs a lower bound on the BFE that is a convex function of $q$.  Lagrange duality is then used to demonstrate that the lower bound is finite under the condition of the theorem.

\subsection{Lifting}
Once model symmetries are detected using coloring passing or an alternative method, they can be encoded into the variational objective by introducing constraints on the marginal distributions, e.g., adding a constraint that all variables in the same super node have equivalent marginals.  This is the approach taken by \cite{bui2014} for lifted variational inference in discrete MRFs.  For us, this leads to the following set of constraints.
\begin{align}
\forall i, j \in \mathfrak{i}, \sum_{k=1}^K q_i^k(x_i) = \sum_{k=1}^K q_j^k(x_j) \label{eq:shared_marg}
\end{align}
If preferred, these constraints could be incorporated into the objective as soft penalty terms to encourage the solution to contain the appropriate symmetries as discovered by color-passing.  However, adding constraints of this form to the objective does not reduce the cost of performing inference in the lifted model.  In order to make lifting tractable, we observe that the following constraints are sufficient for \eqref{eq:shared_marg} to hold.
\begin{align}
\forall i, j \in \mathfrak{i}, \forall_k, q_i^k(x_i) = q_j^k(x_j) \label{eq:shared_comp}
\end{align}
Under the constraint \eqref{eq:shared_comp}, we can simplify the variational objective by accounting for the shared parameters.  As an example of how this works for the Bethe free energy approximation, consider a compressed graph $\mathfrak{G}$ with a set of super variables $\mathfrak{V}$ and a set of super factors $\mathfrak{C}$, each $\mathfrak{i} \in \mathfrak{V}$ and each $\mathfrak{c} \in \mathfrak{C}$ corresponds to $\#(\mathfrak{i})$ variables and $\#(\mathfrak{c})$ factors in the original graph. Variables in the super variable $\mathfrak{i}$ share the same parameterized marginals. Based on this observation, we can simplify the computation of the unlifted BFE by exploiting these symmetries:
\begin{align*}
\sum_{\mathfrak{i} \in \mathfrak{V}} \#(\mathfrak{i}) &\cdot \mathbb{E}_{q_{\mathfrak{i}}(x_{\mathfrak{i}})} \Big[(1 - |nb(\mathfrak{i})|) \log q_{\mathfrak{i}}(X_{\mathfrak{i}})\Big] + \\&\sum_{\mathfrak{c} \in \mathfrak{C}} \#(\mathfrak{c}) \cdot \mathbb{E}_{q_{\mathfrak{c}}^{k}(x_{\mathfrak{c}})} \Big[\log q_\mathfrak{c}(X_\mathfrak{c}) - \log \psi_\mathfrak{c}(X_\mathfrak{c}) \Big].
\end{align*}
Although the optimal value of the variational objective under the constraints \eqref{eq:shared_marg} or \eqref{eq:shared_comp} is always greater than or equal to that of the unconstrained problem, we expect gradient descent on the constrained optimization problem to converge faster and to a better solution as the optimal solution should contain these symmetries. The intuition for this is that that the solutions to the unconstrained optimization problem, i.e., approximate inference in the unlifted model, can include both solutions that do and do not respect the model symmetries.

\subsection{Algorithm}
Given a ground MRF $p$ (possibly conditioned on evidence), we first obtain the variational distrubtion $q^*$ by gradient descent on the variational objective $\mathcal{F}(q)$  w.r.t. the parameters $(w, \eta)$ of mean-field variational mixture $q$,  where the $i$th marginal $q_i^k$ is taken to be a Gaussian distribution for all continuous $i\in V$ and a categorical distribution otherwise.
The lifted variational inference algorithms additionally exploit symmetries by using \eqref{eq:shared_comp} to simplify the objective and only optimize over the variational parameters $\eta_{\mathfrak{i}}$ associated with the super variables ${\mathfrak{i}} \in \mathfrak{V}$; after the optimization procedure, all the original variables contained in each super variable are assigned the same variational marginal parameters/distributions as in \eqref{eq:shared_comp}. 
The expectations in the variational objectives can be approximated in different ways, e.g., sampling, Stein variational methods \cite{liu2016}, etc.  We approximate the expectations using Gaussian quadrature with a fixed number of quadrature points \cite{golub1969}. Once $q^*$ is obtained, given a set of query variables $U$, marginal inference is approximated by $p(x_U) \approx q^*(x_U) = \sum_{k=1}^K w_k \prod_{i\in U} q_i^{*k}(x_i)$, and (marginal) MAP inference is approximated by $\arg \max_{x_U} q^*(x_U)$ via coordinate/gradient ascent.

\subsection{Coarse-to-Fine Lifting}
A common issue in lifting methods is that introducing evidence breaks model symmetries as variables with different evidence values should be considered as different even if they have similar neighborhood structure in the graphical model. This issue is worse when variables can have continuous values: it is unlikely that two otherwise symmetric variables will receive the same exact evidence values.  As a result, even with a small amount of evidence, many of the model symmetries may be destroyed, making lifting less useful.  To counteract this effect, we propose a coarse-to-fine (C2F) approximate lifting method in the variational setting which is based on the assumption that the stationary points of a coarsely compressed graph and a finely compressed graph should be somewhat similar.  A number of coarse-to-fine lifting schemes, which start with coarse approximate symmetries and gradually refine them, have been proposed for discrete MRFs \cite{habeeb2017,gallo2018}. Our approach is aimed specifically at introducing approximate symmetries to handle the above issue with continuous evidence.

Our C2F approximate lifting uses $k$-means clustering to group the continuous evidence values into $s$ clusters, $E_1, \ldots, E_s$. For each cluster $E_i \in \{E_1, \ldots, E_s\}$, we denote the corresponding set of observation nodes as $O_i$. Each observed variable $o \in O_i$ is treated as having the same evidence distribution $b_{E_i}(x_o) = \mathcal{N}(\mu_{E_i}, \sigma_{E_i}^{2})$, where $\mu_{E_i}$ and $\sigma_{E_i}^2$ are the mean and variance of cluster $E_i$. With this formalism, the evidence clustering is coarse when $s$ is small, but we can exploit more approximate symmetries, resulting in a more compressed lifted graph. As $s$ increases, the evidence variables are more finely divided.

To apply this lifting process in variational inference, we interleave the operation of refining the compressed graph and gradient descent. The clustering is initialized with $s=1$ and CP is run until convergence to obtain a coarse compressed graph. Then, we perform gradient descent on the coarse compressed graph with the modified variational method. After a specified number of iterations, we refine the coarse compressed graph by splitting evidence clusters. We employ the $k$-means algorithm to determine the new evidence clusters, after which a refined compressed graph can be obtained through CP. We keep iterating this process until no evidence group can be further split, e.g., when only one value remains or the variance of each cluster is below a specified threshold, and the optimization converges to a stationary point. A precise description of this process can be found in Algorithm~\ref{alg:c2f}.  It is not necessary to rerun CP from the start after each split. We simply assign a new evidence group and a new color and resume CP from its previous stopping point.

\begin{algorithm}[t]
   \caption{Coarse-to-Fine Lifted VI}
   \label{alg:c2f}
\begin{algorithmic}[1]
   \State {\bfseries Input:} A factor graph $G$ with variables $\mathcal{V}$, factors $\mathcal{C}$, evidence $E$ and splitting threshold $\epsilon$
   \State {\bfseries Return:} The model parameters $\eta$ and $w$
   \State $\mathcal{E},\eta, w \gets$ initial clustering of continuous evidence and model parameters respectively
   \State $\mathfrak{G} \gets$ run CP starting from initial colors based on domain/evidence and potential function
   \Repeat
   \State $\eta, w \gets$ run grad. descent on variational obj.
   \For{each $E_i \in \mathcal{E}$}
       \If{$\sigma_{Ei}^2 > \epsilon$}
       \State $\mathcal{E}_i \gets$ Divide $E_i$ in two using $k$-means
       \State $\mathcal{E} \gets (\mathcal{E}\setminus E_i) \cup \mathcal{E}_i$
   \EndIf
   \EndFor
   \State Assign new colors to evidence according to $\mathcal{E}$
   \State $\mathfrak{G} \gets \text{run CP until convergence}$
   \Until{Converge}
\end{algorithmic}
\end{algorithm}

\section{Experiments}
We investigate the performance of the above lifted variational approach on a variety of both real and synthetic models.  We aim to answer the following questions explicitly:
\begin{itemize}
\item[\textbf{Q1:}]{Do the proposed variational approaches yield accurate MAP and marginal inference results?}
\item[\textbf{Q2:}]{Does lifting result in significant speed-ups versus an unlifted variational method?}
\item[\textbf{Q3:}]{Does C2F lifting yield accurate results more quickly for queries with continuous evidence?}
\end{itemize}

To answer these questions, we compare the performance of our variational approach with different entropy approximations (BVI for Bethe approximation and NPVI for the Jensen lower bound) with message-passing algorithms including our own lifted version of Expectation Particle BP (EPBP) \cite{lienart2015} and Gaussian BP (GaBP). To illustrate the generality of the proposed approach, we consider different settings -- Relational Gaussian Models (RGMs) \cite{choi2010}, Relational Kalman Filters (RKFs) \cite{choi2011}, and (Hybrid) Markov Logic Networks (HMLNs) \cite{richardson2006,wang08}. 

For evaluation, we generally report the $\ell_1$ error of MAP predictions and KL divergence $D_{KL}(p_i||q_i)$ averaged across all univariate marginals $i$ in the (ground) conditional MRF. As the models in the RGM and RKF experiments are Gaussian MRFs, their marginal means and variances can be computed exactly by matrix operations. For the HMLN experiments, the exact ground truth can be obtained by direct methods when the number of random variables in the conditional MRF is small. For timing experiments, unless otherwise noted, were performed on a single core of a machine with a 2.2 GHz Intel Core i7-8750H CPU and 16 GB of memory.  Source code is implemented with Python3.6 and is available online \url{github.com/leodd/Lifted-Hybrid-Variational-Inference}.

\subsection{Hybrid MLNs}
We first consider a toy HMLN with known ground truth marginals in order to assess the accuracy of the different variational approaches in the hybrid setting.  Then, we showcase the efficiency of our methods, particularly via lifting, on larger-scale HMLNs of practical interest, comparing against a state-of-the-art sampling baseline. 

\subsubsection{Toy Problem}
We construct a hybrid MLN for a \emph{position} domain: 
\begin{align*}
0.1 &: in(A, Box) \wedge in(B, Box) \rightarrow attractedTo(A, B) \\
0.2 &: \neg attractedTo(A, B) \cdot [pos(A) = p1] +\\
& attractedTo(A, B) \cdot [pos(B) = p2],
\end{align*}
where $A$ and $B$ are different classes of objects in some physics simulation, $Box$ is the class of box instances, and $p1, p2$ are real values corresponding to object positions. The first formula states that object $A$ is attracted to $B$ when they are in the same box; the second states that if $A$ is attracted to $B$, then $A$ is likely at position $p1$, otherwise $B$ is likely at position $p2$. Predicates $in$ and $attractedTo$ have discrete domain $\{0, 1\}$, while $pos$ is real-valued. 

Unlike standard MLNs, where the value of a formula can be computed with logical operations, an HMLN defines continuous operations for hybrid formulas. In the second formula, 
$\alpha = \beta$ is shorthand for the feature function $-(\alpha - \beta)^2$ ; note that the corresponding linear Gaussian potential $\psi(x_1, x_2)=e^{-(x_1 - x_2)^2}$ is \emph{not normalizable}.
The marginals of $pos(A)$ and $pos(B)$ will generally be multimodal, specifically mixtures of Gaussians; with multiple object instances, unimodal variational approximations like mean-field will likely be inaccurate.
\begin{figure}[t]
    \centering
    \includegraphics[scale=0.4]{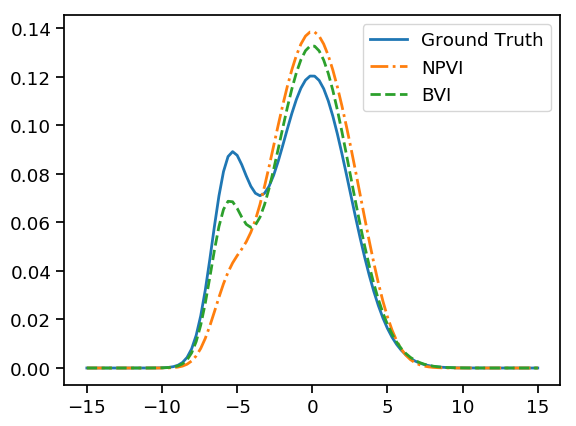}
    \caption{\small Visualization of typical marginal distribution estimates produced by BVI and NPVI on the toy HMLN experiment, using $K=5$. The lifted solutions were similar. }
\label{fig:marg}
\end{figure}
Note that the number of mixture components in the marginal distributions is exponential in the number of joint discrete configurations in the ground MRF, so we consider a small model in which exact inference is still tractable,  generating 2, 3, and 2 instances of the $A$, $B$, and $Box$ classes respectively. The resulting ground MRF contains $16$ discrete 
 random variables yielding marginals with $2^{16}$ mixture components.  We emphasize that the  small model size is only for the purpose of evaluation against brute-force exact inference; our methods can scale to much larger models.


We performed marginal inference on the continuous nodes and report results against ground truth in Table~\ref{tab:mln}, using BVI, NPVI, and their lifted versions (dubbed L-BVI and L-NPVI).  
All methods tend to give improved performance as the number of mixture components ($K$) increases indicating that the number of mixture components is indeed important for accuracy in multimodal settings. However, increasing $K$ generally makes the optimization problem more difficult, requiring more iterations for convergence. 
We note that even though L-BVI reported a lower $\ell_1$ error with $K=1$, the KL-divergence of this was larger than at $K=3$ or $5$, indicating that it converged to a good local optimum for the MAP task but not as good for the marginal inference task. This distinction can be seen more broadly across the two entropy approximations for this problem, as BVI/L-BVI generally gave better fits to the marginals than NPVI/L-NPVI, whereas NPVI/L-NPVI performed better at estimating the marginal modes. See Figure \ref{fig:marg} for an example illustration of marginals produced by the different entropy approximations.

It is also worth noting that {\em lifting seems to act as a regularizer} here:  when the number of mixture components is small, both the lifted versions outperformed their unlifted counterparts, e.g., at $K$=1.  This suggests that lifting may both reduce computational cost ($30\%$ to $40\%$ speedup on this model) and encourage the optimization procedure to end up in better local optima, which 
positively answers \textbf{Q1} and \textbf{Q2}.

\begin{table}[t]
\centering
\caption{Results of variational methods on toy HMLN.}
\scalebox{.8}{
\begin{tabular}{|c|ccc|ccc|}
\hline
Algorithm & \multicolumn{3}{c|}{Average KL-Divergence} \\
 & $K = 1 $ & $K = 3$ & $K = 5$ \\
\hline
BVI & $ 0.513 \pm 0.947$ & $0.009 \pm 0.006$ & $0.009 \pm 0.007$\\

L-BVI 
&$0.039 \pm 1 \mathrm{e}-5$&  $\mathbf{0.004 \pm 0.001}$ & $0.004\pm 0.002$\\

NPVI
& $4.586\pm 2.977$ & $0.026\pm 0.007$& $0.022\pm0.003$\\

L-NPVI 
& $1.978\pm 3.878$&  $0.038\pm 0.002$& $0.039\pm 0.001$\\
\hline

\hline
Algorithm & \multicolumn{3}{c|}{Average $\ell_1$-error} \\
 & $K = 1 $ & $K = 3$ & $K = 5$ \\
\hline
BVI
&$0.227\pm0.440$ & $0.059\pm 0.026$ & $0.049\pm 0.035$\\

L-BVI 
&$0.007\pm 1\mathrm{e}-4$ & $0.049\pm 0.027$ & $0.020\pm 0.012$\\

NPVI
&$1.641\pm 1.106$ & $\mathbf{0.007\pm 4 \mathrm{e}-4}$ & $0.012\pm 0.009$ \\

L-NPVI 
&$0.671\pm 1.327$ & $0.012\pm 0.006$ & $0.012\pm 0.006$\\

\hline
\end{tabular}
}
\label{tab:mln}
\end{table}

\subsubsection{Larger-scale Problems}
Next, we consider two larger scale HMLNs of practical interest. The \textbf{Paper Popularity} HMLN domain is determined by the following formulas.
\begin{align*}
0.3 &: PaperPopularity(p) = 1.0 \\
0.5 &: SameSession(t1,t2) \cdot \\
 & [TopicPopularity(t1) = TopicPopularity(t2)] \\
0.5 &: In(p,t) \cdot \\
& [PaperPopularity(p) = TopicPopularity(t)],
\end{align*}
where $PaperPopularity(p)$ and $TopicPopularity(t)$ are continuous variables in $[0, 10]$, indicating the popularity of paper and topic. $SameSession(t1, t2)$ and $In(p, t)$ are Boolean variables, indicating if two topics are in the same session, and if a paper $p$ belongs to a topic $t$, respectively. The first formula specifies a prior on paper popularity. The second clause states that two topics tend to have the same popularity if they are in the same session, and the third clause states that if a paper is in a topic, then the popularity of the topic and the paper are likely to be the same. We instantiate 300 paper instances and 10 topic instances, which results in $3,400$ variable nodes and $3,390$ factor nodes in the grounded MRF. We generated random evidence for the model, where $70\%$ of the papers and $70\%$ of the topics were assigned a popularity from a uniform distribution $U(0, 10)$. For $70\%$ of the papers $p$ and all topics $t$, we assign $In(p,t')$ and $SameSession(t,t')$ for all possible $t' \in Topic$ using a Bernoulli distribution ($p=0.5$).
\begin{figure}[t]
    \centering
    \begin{subfigure}{0.4\textwidth}
      \includegraphics[width=1.0\linewidth]{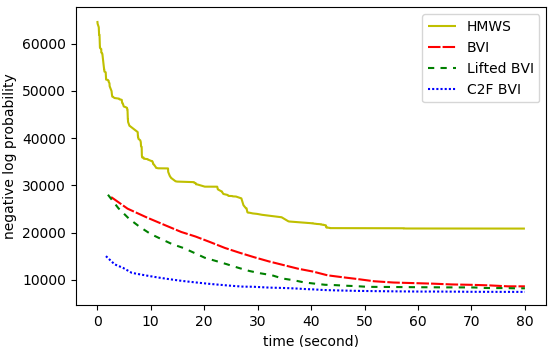}
      \caption{Paper Popularity}
      \label{fig:sfig1}
    \end{subfigure}
    \hfill \hfill
    \begin{subfigure}{0.4\textwidth}
      \includegraphics[width=1.0\linewidth]{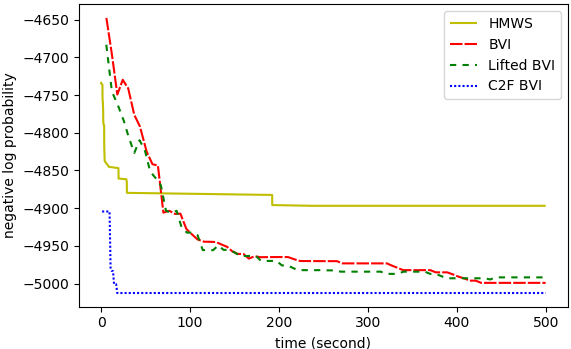}
      \caption{Robot Mapping}
      \label{fig:sfig2}
    \end{subfigure}
    \caption{\small Comparison of the negative log probability of the approximate MAP assignment versus running time.} 
    \label{fig:time-log-paper-popularity}
\end{figure}

The \textbf{Robot Mapping} HMLN domain contains $3$ discrete relational variables, $2$ continuous relational variables, and $10$ formulas, as described in the Alchemy tutorial \cite{wang08}. The instances and evidences are from real world robot scanning data, which result in a grounded MRF with $1591$ random variables and $3182$ factors.

\textbf{Results:}
We performed MAP inference with the variational methods and evaluated them against Hybrid MaxWalkSAT (HMWS) \cite{wang08}.  Each method is evaluated by computing the energy $E(\hat x) = -\log(\prod_{c \in \mathcal{C}} \psi_c(\hat{x_c}))$, essentially the negative log probability, of the approximate MAP configuration that it produces. For HMWS, we set the greedy probability to $0.7$, the standard deviation of the Gaussian noise to $0.3$, and disabled re-running for fair comparison. For the variational methods, we used ADAM with learning rate $0.2$ for optimization. 

As can be seen in Figure \ref{fig:time-log-paper-popularity}, in both domains, the MAP assignment produced by the variational methods is significantly better than the one by HMWS, providing evidence for \textbf{Q1}.  
In addition, given the amount of continuous evidence, there is not a significant performance difference between BVI and Lifted BVI.  However, C2F BVI takes less significantly less time to converge to a good solution than both BVI and Lifted BVI, providing strong evidence for \textbf{Q3}, namely that C2F results in better accuracy more quickly. 

\subsection{Relational Gaussian Model}
In this experiment, we performed approximate inference on a Relational Gaussian Model (RGM) with the recession domain from \cite{csekeheskesjair2011}. The  RGM has three relational atoms $Market(S)$, $Loss(S, B)$, $Revenue(B)$ and one random variable $Recession$, where $S$ and $B$ denote two sets of instances, the categories of market and banks respectively. For testing, we generated $100$ Market and $5$ Bank instances, and used the ground graph as input.   To assess the impact of lifting and C2F, we randomly chose $20\%$ of the variables in the model, assigned them a value uniformly randomly from $[-30, 30]$, and then performed conditional MAP and marginal inference. 

\begin{table}[t]
     \caption{Evaluation of various methods on RGM.}
\centering
\scalebox{0.8}{
\begin{tabular}{|c|c|c|}
\hline
\multicolumn{1}{|c|}{Algorithm} & \multicolumn{1}{c|}{Avg. $\ell_1$ Error} & \multicolumn{1}{c|}{Avg. KL-Divergence} \\
\hline
EPBP & $5.17\mathrm{e}{-2} \pm 4.25\mathrm{e}{-2}$ & $0.473 \pm 0.246$\\
GaBP & $\mathbf{2.26\mathrm{e}{-6} \pm 1.14\mathrm{e}{-6}}$ & $\mathbf{3.32\mathrm{e}{-9} \pm 1.12\mathrm{e}{-9}}$\\
BVI & $6.48\mathrm{e}{-5} \pm 2.57\mathrm{e}{-4}$ & $4.95\mathrm{e}{-3} \pm 5.13\mathrm{e}{-2}$\\
L-BVI & $5.77\mathrm{e}{-5} \pm 2.47\mathrm{e}{-4}$ & $4.95\mathrm{e}{-3} \pm 5.13\mathrm{e}{-2}$\\
C2F-BVI & $7.92\mathrm{e}{-4} \pm 2.31\mathrm{e}{-3}$ & $4.95\mathrm{e}{-3} \pm 5.13\mathrm{e}{-2}$\\
NPVI & $3.22\mathrm{e}{-5} \pm 2.89\mathrm{e}{-5}$ & $5.14\mathrm{e}{-4} \pm 4.14\mathrm{e}{-5}$\\
L-NPVI & $3.29\mathrm{e}{-5} \pm 5.58\mathrm{e}{-5}$ & $5.14\mathrm{e}{-4} \pm 4.14\mathrm{e}{-5}$\\
\hline
\end{tabular}}
\label{tab:rgm}
\end{table}

\begin{figure}
    \centering
    \includegraphics[scale=0.48]{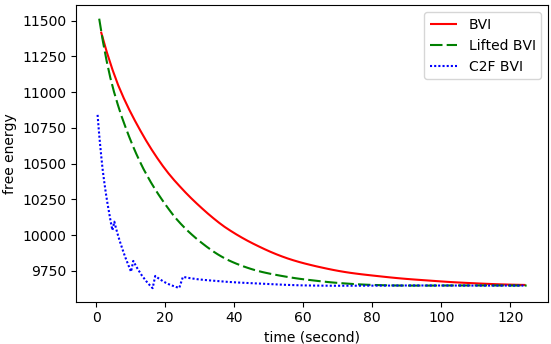}
    \caption{\small Comparison of rate of convergence between BVI, L-BVI, and C2F-BVI with $20\%$ evidence, on RGM.}
\label{fig:time-log-rgm}
\end{figure}

Figure \ref{fig:time-log-rgm} plots the value of the free energy, i.e., the variational objective, versus time for BVI, L-BVI, and C2F-BVI.  All three methods used the Adam optimizer with learning rate $0.2, \beta_1 = 0.9, \beta_2 = 0.999$. The plot shows that C2F-BVI converges faster than L-BVI, which is in turn faster than BVI. Note that the sawtooth shape of C2F-BVI is a result of evidence splitting.  This provides evidence that lifting and C2F do reduce the time cost of inference, answering \textbf{Q2} and \textbf{Q3} in the affirmative.  

To assess the accuracy of the variational methods, we randomly chose $5\%$ to $20\%$ of the random variables and generated evidence values as in the previous task. We randomly generated five evidence settings and evaluated all variational methods with the same setup as above as well as EPBP with 20 sampling points and GaBP. All algorithms were run until convergence and compared against the ground truth. As Table~\ref{tab:rgm} shows, on this simple unimodal model, all variational methods have very low error and low KL-Divergence, while EPBP has higher error introduced by the sampling procedure, providing evidence for \textbf{Q1} that the variational approach is accurate in this case.  In this unimodal case, NPVI appears to outperform BVI.  We note that, as a general observation, NPVI tends to do better than BVI at estimating the mode of the distribution, but in multimodal settings, tends to result in a higher KL-divergence than BVI.

\subsection{Relational Kalman Filtering}
To further investigate \textbf{Q1}, we performed an experiment with Relational Kalman Filters (RKFs). A standard Kalman filter (KF) models the transition of a dynamic system with $x_{t+1} = A x_t + w$ with noise $o_t = C x_t + v$, where $A$ denotes the transition matrix and $C$ represents the observation matrix. A key assumption in KF is that the transition and noise follow a normal distribution, i.e., $w \sim \mathcal{N}(0, Q)$, and $v\sim \mathcal{N}(0, R)$, for covariance matrices $Q$ and $R$. The RKF model defines a lifted KF, i.e. similar state variables and similar observation nodes share the same transition and observation model.

In this experiment, we use extracted groundwater level data from the Republican River Compact Association model \cite{mckusick2003}. The data set contains a record over 850 months of the water level data for 3,420 wells. We followed the same data preprocessing steps as in \cite{choi2015}, where wells in the same area are grouped together and are assumed to share same transition and observation0 model. We test our algorithms on two different structure settings, a tree-structured model and a model with cycles. For the tree-structured model, we define the matrices $A = \alpha \cdot I$, $C = I$, $Q = \beta \cdot I$, $R = \gamma \cdot I$ where $\alpha \sim U(0.5, 1)$, $\beta \sim U(5, 10)$, $\gamma \sim U(1, 5)$ and $I$ is the identity matrix. For general model, we select $A = \alpha \cdot I + 0.01$ and $Q = \beta \cdot J$, where $J$ is the matrix of ones, and other matrices are as before. We chose 20 months of record as the observation of the model. Note that the model defined has a linear Gaussian potential $\exp((x_{t+1} - A x_t)^2/\sigma^2)$ with the number of state variables as its dimension, which makes inference challenging. For simplicity, we expressed the model as a product pairwise potentials.

We compared our variational methods and EPBP against GaBP. The variational methods used Adam optimizer with learning rate $0.2$; EPBP used $20$ particle points. Table~\ref{tab:rkf} reports the resulting avg $\ell_1$ difference and KL-divergence of the MAP estimate of the last time step nodes against GaBP.  The variational methods are quite accurate in this model, owing mostly to the unimodality of the marginals, providing evidence in support of \textbf{Q1} for this setting, even when the graph contains cycles. The variational methods also obtained better KL-divergence than EPBP, and the KL-divergence of all methods would likely further reduce with additional mixture components/particles.

Significant speed-ups are obtained by the lifted methods with a small improvement in the case of C2F.  Note that the times are for the relative comparison of the lifting, and while they do indicate significant performance improvements from lifting, answering \textbf{Q2} affirmatively, they should not be used to compare EPBP and BVI as no implementations were optimized for performance in this case.  As the variational methods can also be efficiently implemented in Tensorflow, we also provide timing experiments in a high performance setting for BVI/NPVI and their lifted versions. For BVI/L-BVI, the TensorFlow implementation took $107.8/5.6s$ to run on the tree model and $16.3/12.6s$ on the cycle model; NPVI/L-NPVI took $105.7/5.11s$ and $15.7/12.8s$ on the tree and cycle models respectively, and gave identical performance to BVI/L-BVI.

\begin{table}[t]
\caption{\small Accuracy of lifted inference methods for RKFs against GaBP.}
\centering
\renewcommand{\arraystretch}{1.2}
\scalebox{.8}{
\begin{tabular}{|c|c|c|c|}
\hline
& \multicolumn{1}{c|}{Algorithm} & \multicolumn{1}{c|}{$\ell_1$ Diff.(GaBP)} & \multicolumn{1}{c|}{time (s)}\\
\hline
\parbox[t]{2mm}{\multirow{6}{*}{\rotatebox[origin=c]{90}{Tree}}}
& EPBP & $0.18 \pm 0.15$ & $233.6$\\
& BVI & $2.86\mathrm{e}{-5} \pm 3.71\mathrm{e}{-5}$ & $577.2$\\
& L-BVI & $\mathbf{1.88\mathrm{e}{-5} \pm 1.85\mathrm{e}{-5}}$ & $23.0$\\
& C2F-BVI & $1.64\mathrm{e}{-5} \pm 1.80\mathrm{e}{-5}$ & $\mathbf{22.8}$\\
\hline
\parbox[t]{2mm}{\multirow{6}{*}{\rotatebox[origin=c]{90}{Cycle}}}
& EPBP & $0.34 \pm 0.41$ & $149.2$\\
& BVI & $\mathbf{3.26\mathrm{e}{-5} \pm 5.08\mathrm{e}{-5}}$ & $222.1$\\
& L-BVI & $4.65\mathrm{e}{-5} \pm 5.78\mathrm{e}{-5}$ & $151.0$\\
& C2F-BVI & $2.16\mathrm{e}{-4} \pm 4.38\mathrm{e}{-4}$ & $\mathbf{140.8}$\\
\hline
\end{tabular}}
\label{tab:rkf}
\end{table}

\section{Discussion}

We presented distribution-independent, model-agnostic hybrid lifted inference algorithm that makes minimal assumptions on the underlying distribution.  In addition, we presented a simple coarse-to-fine approach for handling symmetry breaking as a result of continuous evidence. We showed experimentally that the lifted and the coarse-to-fine variational methods compare favorably in terms of accuracy against exact and particle based methods for MAP and marginal inference tasks and can yield speed-ups over their non-lifted counterparts that range from moderate to significant, depending on the amount of evidence and the distribution of the evidence values.  Finally, we provided a sufficient condition under which the BFE over the marginal polytope is bounded from below and showing that the BFE approximations yields a nontrivial approximation to the partition function in these cases.

\bibliography{reference}

\begin{thebibliography}{}

\bibitem[\protect\citeauthoryear{Ahmadi, Kersting, and
  Sanner}{2011}]{ahmadi2011}
Ahmadi, B.; Kersting, K.; and Sanner, S.
\newblock 2011.
\newblock Multi-evidence lifted message passing, with application to
  {P}age{R}ank and the {K}alman filter.
\newblock In {\em Proceedings of the 20th International Joint Conference on
  Artificial Intelligence}.

\bibitem[\protect\citeauthoryear{Bui, Huynh, and Sontag}{2014}]{bui2014}
Bui, H.~H.; Huynh, T.~N.; and Sontag, D.
\newblock 2014.
\newblock Lifted tree-reweighted variational inference.
\newblock In {\em Proceedings of the Thirtieth Conference on Uncertainty in
  Artificial Intelligence (UAI)}.

\bibitem[\protect\citeauthoryear{Choi, Amir, and Hill}{2010}]{choi2010}
Choi, J.; Amir, E.; and Hill, D.~J.
\newblock 2010.
\newblock Lifted inference for relational continuous models.
\newblock In {\em Proceedings of the Twenty-Sixth Conference on Uncertainty in
  Artificial Intelligence (UAI)}.

\bibitem[\protect\citeauthoryear{Choi and Amir}{2012}]{choi2012}
Choi, J., and Amir, E.
\newblock 2012.
\newblock Lifted relational variational inference.
\newblock In {\em Proceedings of the Twenty-Eighth Conference on Uncertainty in
  Artificial Intelligence (UAI)}.

\bibitem[\protect\citeauthoryear{Choi \bgroup et al\mbox.\egroup
  }{2015}]{choi2015}
Choi, J.; Amir, E.; Xu, T.; and Valocchi, A.~J.
\newblock 2015.
\newblock Learning relational {K}alman filtering.
\newblock In {\em Twenty-Ninth AAAI Conference on Artificial Intelligence
  (AAAI)}.

\bibitem[\protect\citeauthoryear{Choi, Guzman-Rivera, and
  Amir}{2011}]{choi2011}
Choi, J.; Guzman-Rivera, A.; and Amir, E.
\newblock 2011.
\newblock Lifted relational {K}alman filtering.
\newblock In {\em Twenty-Fifth AAAI Conference on Artificial Intelligence
  (AAAI)}.

\bibitem[\protect\citeauthoryear{Cseke and Heskes}{2011}]{csekeheskesjair2011}
Cseke, B., and Heskes, T.
\newblock 2011.
\newblock Properties of {B}ethe free energies and message passing in {G}aussian
  models.
\newblock {\em Journal of Artificial Intelligence Research}  1--24.

\bibitem[\protect\citeauthoryear{Gallo and Ihler}{2018}]{gallo2018}
Gallo, N., and Ihler, A.
\newblock 2018.
\newblock Lifted generalized dual decomposition.
\newblock In {\em Thirty-Second AAAI Conference on Artificial Intelligence
  (AAAI)}.

\bibitem[\protect\citeauthoryear{Gershman, Hoffman, and
  Blei}{2012}]{gershman2012}
Gershman, S.~J.; Hoffman, M.~D.; and Blei, D.~M.
\newblock 2012.
\newblock Nonparametric variational inference.
\newblock In {\em Proceedings of the 29th International Conference on Machine
  Learning (ICML)},  235--242.

\bibitem[\protect\citeauthoryear{Golub and Welsch}{1969}]{golub1969}
Golub, G.~H., and Welsch, J.~H.
\newblock 1969.
\newblock Calculation of {G}auss quadrature rules.
\newblock {\em Mathematics of computation} 23(106):221--230.

\bibitem[\protect\citeauthoryear{Habeeb \bgroup et al\mbox.\egroup
  }{2017}]{habeeb2017}
Habeeb, H.; Anand, A.; Mausam, M.; and Singla, P.
\newblock 2017.
\newblock Coarse-to-fine lifted map inference in computer vision.
\newblock In {\em Proceedings of the 26th International Joint Conference on
  Artificial Intelligence (IJCAI)}.

\bibitem[\protect\citeauthoryear{Jaakkola and Jordan}{1998}]{jaakkola1998}
Jaakkola, T.~S., and Jordan, M.~I.
\newblock 1998.
\newblock Improving the mean field approximation via the use of mixture
  distributions.
\newblock In {\em Learning in graphical models}. Springer.
\newblock  163--173.

\bibitem[\protect\citeauthoryear{Kersting, Ahmadi, and
  Natarajan}{2009}]{kersting2009}
Kersting, K.; Ahmadi, B.; and Natarajan, S.
\newblock 2009.
\newblock Counting belief propagation.
\newblock In {\em Proceedings of the Twenty-Fifth Conference on Uncertainty in
  Artificial Intelligence (UAI)}.

\bibitem[\protect\citeauthoryear{Lienart, Teh, and Doucet}{2015}]{lienart2015}
Lienart, T.; Teh, Y.~W.; and Doucet, A.
\newblock 2015.
\newblock Expectation particle belief propagation.
\newblock In {\em NeurIPS}.

\bibitem[\protect\citeauthoryear{Liu, Lee, and Jordan}{2016}]{liu2016}
Liu, Q.; Lee, J.; and Jordan, M.
\newblock 2016.
\newblock A kernelized {S}tein discrepancy for goodness-of-fit tests.
\newblock In {\em International Conference on Machine Learning (ICML)},
  276--284.

\bibitem[\protect\citeauthoryear{McKusick}{2003}]{mckusick2003}
McKusick, V.
\newblock 2003.
\newblock Final report for the special master with certificate of adoption of
  {RRCA} groundwater model.
\newblock {\em State of Kansas v. State of Nebraska and State of Colorado, in
  the Supreme Court of the United States}.

\bibitem[\protect\citeauthoryear{Richardson and
  Domingos}{2006}]{richardson2006}
Richardson, M., and Domingos, P.
\newblock 2006.
\newblock Markov logic networks.
\newblock {\em Machine Learning} 62.

\bibitem[\protect\citeauthoryear{Ruozzi}{2012}]{nips2012}
Ruozzi, N.
\newblock 2012.
\newblock The {B}ethe partition function of log-supermodular graphical models.
\newblock In {\em Neural Information Processing Systems (NeurIPS)}.

\bibitem[\protect\citeauthoryear{Ruozzi}{2013}]{uai2013}
Ruozzi, N.
\newblock 2013.
\newblock Beyond log-supermodularity: Lower bounds and the {B}ethe partition
  function.
\newblock In {\em Proceedings of the Twenty-Ninth Conference Annual Conference
  on Uncertainty in Artificial Intelligence (UAI)}.

\bibitem[\protect\citeauthoryear{Ruozzi}{2017}]{ruozzi2017}
Ruozzi, N.
\newblock 2017.
\newblock A lower bound on the partition function of attractive graphical
  models in the continuous case.
\newblock In {\em Artificial Intelligence and Statistics (AISTATS)}.

\bibitem[\protect\citeauthoryear{Singla and Domingos}{2008}]{singla2008}
Singla, P., and Domingos, P.
\newblock 2008.
\newblock Lifted first-order belief propagation.
\newblock In {\em Twenty-Second AAAI Conference on Artificial Intelligence
  (AAAI)}.

\bibitem[\protect\citeauthoryear{Wang and Domingos}{2008}]{wang08}
Wang, J., and Domingos, P.
\newblock 2008.
\newblock Hybrid markov logic networks.
\newblock In {\em Twenty-Second AAAI Conference on Artificial Intelligence
  (AAAI)}.

\bibitem[\protect\citeauthoryear{Welling and Teh}{2001}]{welling2001}
Welling, M., and Teh, Y.~W.
\newblock 2001.
\newblock Belief optimization for binary networks: A stable alternative to
  loopy belief propagation.
\newblock In {\em Proceedings of the Seventeenth Conference on Uncertainty in
  Artificial Intelligence (UAI)}.

\bibitem[\protect\citeauthoryear{Yedidia, Freeman, and
  Weiss}{2001}]{yedidia2001}
Yedidia, J.~S.; Freeman, W.~T.; and Weiss, Y.
\newblock 2001.
\newblock Bethe free energy, {K}ikuchi approximations, and belief propagation
  algorithms.
\newblock {\em Neural Information Processing Systems (NeurIPS)}.

\bibitem[\protect\citeauthoryear{Yedidia, Freeman, and Weiss}{2005}]{yedweiss}
Yedidia, J.~S.; Freeman, W.~T.; and Weiss, Y.
\newblock 2005.
\newblock Constructing free-energy approximations and generalized belief
  propagation algorithms.
\newblock {\em Information Theory, IEEE Transactions on} 51(7):2282 -- 2312.

\end{thebibliography}
\bibliographystyle{aaai}

\end{document}